\documentclass[letterpaper, 10 pt, conference]{ieeeconf}  

\IEEEoverridecommandlockouts                              
\usepackage{epsfig} 
\usepackage{amsmath} 
\usepackage{amssymb}  
\usepackage[utf8]{inputenc} 
\usepackage[T1]{fontenc}
\usepackage{url}
\usepackage{hyperref}
\usepackage{tabularx}
\usepackage{booktabs}

\usepackage[disable]{todonotes}

\usepackage{graphicx,float}
\graphicspath{{figures/}}

\usepackage{tikz}
\usepackage{pgfplots}
\usetikzlibrary{intersections}
\usepackage{bm}
\usetikzlibrary{positioning}
\usepgfplotslibrary{fillbetween}
\pgfplotsset{compat=1.5}
\usetikzlibrary{pgfplots.groupplots}
\usetikzlibrary{shapes.geometric,backgrounds}

\usepackage{tikz}
\usepackage{xifthen}

\pgfdeclarelayer{bg}    
\pgfsetlayers{bg,main}  
\definecolor{beige}{RGB}{245, 245, 220}

\definecolor{darkgrey}{RGB}{75, 75, 75}
\definecolor{lightgrey}{RGB}{250, 250, 250}

\usetikzlibrary{calc, arrows, fit, positioning, patterns, decorations.pathreplacing, shapes}
\tikzstyle{dash} = [dashed, -latex,>=latex]
\tikzstyle{line} = [draw, -latex,>=latex]
\tikzstyle{smallbox} = [draw, minimum size=5.0mm]
\tikzstyle{box} = [draw, minimum size=7.0mm]
\tikzstyle{bigbox} = [draw, minimum size=10.0mm]
\tikzstyle{rectangle} = [draw, minimum width=10.0mm, minimum height=20.0mm]
\tikzstyle{switch} = [trapezium, trapezium angle=120, draw, rotate=90,  inner ysep=5pt, outer sep=5pt,
minimum height=7mm, minimum width=7mm]
\tikzstyle{roundbox} = [draw, circle, inner sep=0pt, minimum size=3mm]
\tikzstyle{clamped} = [draw, fill=darkgrey, minimum size=0.15cm]
\tikzstyle{msgcircle} = [shape=circle, draw, inner sep=0pt, minimum size=4mm, fill=white, font=\scriptsize]
\tikzstyle{darkmsgcircle} = [shape=circle, draw, inner sep=0pt, minimum size=4mm, fill=darkgrey, text=white, font=\scriptsize]
\tikzstyle{redmsgcircle} = [shape=circle, draw=red, inner sep=0pt, minimum size=4mm, text=red, font=\scriptsize]
\tikzstyle{reddarkmsgcircle} = [shape=circle, draw=red, inner sep=0pt, minimum size=4mm, fill=red, text=white, font=\scriptsize]
\tikzstyle{msgdoublecircle} = [shape=circle, double, double distance=1.5pt, draw, inner sep=0pt, minimum size=5mm, fill=white]
\tikzstyle{darkmsgdoublecircle} = [shape=circle, double, double distance=1.5pt, draw, inner sep=0pt, minimum size=5mm, fill=darkgrey, text=white, font=\bfseries]

\newcommand{\msg}[6]{
      \ifthenelse{\isin{#1}{left} \AND \isin{#2}{down}}{
            \coordinate (anchor) at ($({#3})!{#5}!({#4})$);
            \node[xshift=-6.0mm] at (anchor) {#6};
            \node[xshift=-1.0mm] at (anchor) {$\downarrow$};
      }{}
      \ifthenelse{\isin{#1}{right} \AND \isin{#2}{down}}{
            \coordinate (anchor) at ($({#3})!{#5}!({#4})$);
            \node[xshift=6.0mm] at (anchor) {#6};
            \node[xshift=1.0mm] at (anchor) {$\downarrow$};
      }{}

      \ifthenelse{\isin{#1}{down} \AND \isin{#2}{right}}{
            \coordinate (anchor) at ($({#3})!{#5}!({#4})$);
            \node[ yshift=-4.0mm] at (anchor) {#6};
            \node[yshift=-1.0mm] at (anchor) {$\rightarrow$};
      }{}
      \ifthenelse{\isin{#1}{up} \AND \isin{#2}{right}}{
            \coordinate (anchor) at ($({#3})!{#5}!({#4})$);
            \node[ yshift=4.0mm] at (anchor) {#6};
            \node[yshift=1.0mm] at (anchor) {$\rightarrow$};
      }{}

      \ifthenelse{\isin{#1}{down} \AND \isin{#2}{left}}{
            \coordinate (anchor) at ($({#3})!{#5}!({#4})$);
            \node[ yshift=-4.0mm] at (anchor) {#6};
            \node[yshift=-1.0mm] at (anchor) {$\leftarrow$};
      }{}
      \ifthenelse{\isin{#1}{up} \AND \isin{#2}{left}}{
            \coordinate (anchor) at ($({#3})!{#5}!({#4})$);
            \node[ yshift=4.0mm] at (anchor) {#6};
            \node[yshift=1.0mm] at (anchor) {$\leftarrow$};
      }{}

      \ifthenelse{\isin{#1}{left} \AND \isin{#2}{up}}{
            \coordinate (anchor) at ($({#3})!{#5}!({#4})$);
            \node[ xshift=-6.0mm] at (anchor) {#6};
            \node[xshift=-1.0mm] at (anchor) {$\uparrow$};
      }{}
      \ifthenelse{\isin{#1}{right} \AND \isin{#2}{up}}{
            \coordinate (anchor) at ($({#3})!{#5}!({#4})$);
            \node[ xshift=6.0mm] at (anchor) {#6};
            \node[xshift=1.0mm] at (anchor) {$\uparrow$};
      }{}
}

\newcommand{\msgcircle}[6]{
      \ifthenelse{\isin{#1}{left} \AND \isin{#2}{down}}{
            \coordinate (anchor) at ($({#3})!{#5}!({#4})$);
            \node[msgcircle,xshift=-5.0mm] at (anchor) {#6};
            \node[xshift=-1.5mm] at (anchor) {$\downarrow$};
      }{}
      \ifthenelse{\isin{#1}{right} \AND \isin{#2}{down}}{
            \coordinate (anchor) at ($({#3})!{#5}!({#4})$);
            \node[msgcircle,xshift=5.0mm] at (anchor) {#6};
            \node[xshift=1.5mm] at (anchor) {$\downarrow$};
      }{}

      \ifthenelse{\isin{#1}{down} \AND \isin{#2}{right}}{
            \coordinate (anchor) at ($({#3})!{#5}!({#4})$);
            \node[msgcircle, yshift=-5.0mm] at (anchor) {#6};
            \node[yshift=-2.0mm] at (anchor) {$\rightarrow$};
      }{}
      \ifthenelse{\isin{#1}{up} \AND \isin{#2}{right}}{
            \coordinate (anchor) at ($({#3})!{#5}!({#4})$);
            \node[msgcircle, yshift=5.0mm] at (anchor) {#6};
            \node[yshift=2.0mm] at (anchor) {$\rightarrow$};
      }{}

      \ifthenelse{\isin{#1}{down} \AND \isin{#2}{left}}{
            \coordinate (anchor) at ($({#3})!{#5}!({#4})$);
            \node[msgcircle, yshift=-5.0mm] at (anchor) {#6};
            \node[yshift=-2.0mm] at (anchor) {$\leftarrow$};
      }{}
      \ifthenelse{\isin{#1}{up} \AND \isin{#2}{left}}{
            \coordinate (anchor) at ($({#3})!{#5}!({#4})$);
            \node[msgcircle, yshift=5.0mm] at (anchor) {#6};
            \node[yshift=2.0mm] at (anchor) {$\leftarrow$};
      }{}

      \ifthenelse{\isin{#1}{left} \AND \isin{#2}{up}}{
            \coordinate (anchor) at ($({#3})!{#5}!({#4})$);
            \node[msgcircle, xshift=-5.0mm] at (anchor) {#6};
            \node[xshift=-1.5mm] at (anchor) {$\uparrow$};
      }{}
      \ifthenelse{\isin{#1}{right} \AND \isin{#2}{up}}{
            \coordinate (anchor) at ($({#3})!{#5}!({#4})$);
            \node[msgcircle, xshift=5.0mm] at (anchor) {#6};
            \node[xshift=1.5mm] at (anchor) {$\uparrow$};
      }{}
}

\newcommand{\darkmsg}[6]{
      \ifthenelse{\isin{#1}{left} \AND \isin{#2}{down}}{
            \coordinate (anchor) at ($({#3})!{#5}!({#4})$);
            \node[darkmsgcircle, xshift=-5mm] at (anchor) {#6};
            \node[xshift=-1.5mm] at (anchor) {$\downarrow$};
      }{}
      \ifthenelse{\isin{#1}{right} \AND \isin{#2}{down}}{
            \coordinate (anchor) at ($({#3})!{#5}!({#4})$);
            \node[darkmsgcircle, xshift=5mm] at (anchor) {#6};
            \node[xshift=1.5mm] at (anchor) {$\downarrow$};
      }{}

      \ifthenelse{\isin{#1}{down} \AND \isin{#2}{right}}{
            \coordinate (anchor) at ($({#3})!{#5}!({#4})$);
            \node[darkmsgcircle, yshift=-5.0mm] at (anchor) {#6};
            \node[yshift=-2.0mm] at (anchor) {$\rightarrow$};
      }{}
      \ifthenelse{\isin{#1}{up} \AND \isin{#2}{right}}{
            \coordinate (anchor) at ($({#3})!{#5}!({#4})$);
            \node[darkmsgcircle, yshift=5.0mm] at (anchor) {#6};
            \node[yshift=2.0mm] at (anchor) {$\rightarrow$};
      }{}

      \ifthenelse{\isin{#1}{down} \AND \isin{#2}{left}}{
            \coordinate (anchor) at ($({#3})!{#5}!({#4})$);
            \node[darkmsgcircle, yshift=-5.0mm] at (anchor) {#6};
            \node[yshift=-2.0mm] at (anchor) {$\leftarrow$};
      }{}
      \ifthenelse{\isin{#1}{up} \AND \isin{#2}{left}}{
            \coordinate (anchor) at ($({#3})!{#5}!({#4})$);
            \node[darkmsgcircle, yshift=5.0mm] at (anchor) {#6};
            \node[yshift=2.0mm] at (anchor) {$\leftarrow$};
      }{}

      \ifthenelse{\isin{#1}{left} \AND \isin{#2}{up}}{
            \coordinate (anchor) at ($({#3})!{#5}!({#4})$);
            \node[darkmsgcircle, xshift=-5.0mm] at (anchor) {#6};
            \node[xshift=-1.5mm] at (anchor) {$\uparrow$};
      }{}
      \ifthenelse{\isin{#1}{right} \AND \isin{#2}{up}}{
            \coordinate (anchor) at ($({#3})!{#5}!({#4})$);
            \node[darkmsgcircle, xshift=5.0mm] at (anchor) {#6};
            \node[xshift=1.5mm] at (anchor) {$\uparrow$};
      }{}
}

\newcommand{\redbackmsg}[6]{
      \ifthenelse{\isin{#1}{left} \AND \isin{#2}{down}}{
            \coordinate (anchor) at ($({#3})!{#5}!({#4})$);
            \node[reddarkmsgcircle, xshift=-5mm] at (anchor) {#6};
            \node[xshift=-1.5mm] at (anchor) {$\downarrow$};
      }{}
      \ifthenelse{\isin{#1}{right} \AND \isin{#2}{down}}{
            \coordinate (anchor) at ($({#3})!{#5}!({#4})$);
            \node[reddarkmsgcircle, xshift=5mm] at (anchor) {#6};
            \node[xshift=1.5mm] at (anchor) {$\downarrow$};
      }{}

      \ifthenelse{\isin{#1}{down} \AND \isin{#2}{right}}{
            \coordinate (anchor) at ($({#3})!{#5}!({#4})$);
            \node[reddarkmsgcircle, yshift=-5.0mm] at (anchor) {#6};
            \node[yshift=-2.0mm] at (anchor) {$\rightarrow$};
      }{}
      \ifthenelse{\isin{#1}{up} \AND \isin{#2}{right}}{
            \coordinate (anchor) at ($({#3})!{#5}!({#4})$);
            \node[reddarkmsgcircle, yshift=5.0mm] at (anchor) {#6};
            \node[yshift=2.0mm] at (anchor) {$\rightarrow$};
      }{}

      \ifthenelse{\isin{#1}{down} \AND \isin{#2}{left}}{
            \coordinate (anchor) at ($({#3})!{#5}!({#4})$);
            \node[reddarkmsgcircle, yshift=-5.0mm] at (anchor) {#6};
            \node[yshift=-2.0mm] at (anchor) {$\leftarrow$};
      }{}
      \ifthenelse{\isin{#1}{up} \AND \isin{#2}{left}}{
            \coordinate (anchor) at ($({#3})!{#5}!({#4})$);
            \node[reddarkmsgcircle, yshift=5.0mm] at (anchor) {#6};
            \node[yshift=2.0mm] at (anchor) {$\leftarrow$};
      }{}

      \ifthenelse{\isin{#1}{left} \AND \isin{#2}{up}}{
            \coordinate (anchor) at ($({#3})!{#5}!({#4})$);
            \node[reddarkmsgcircle, xshift=-5.0mm] at (anchor) {#6};
            \node[xshift=-1.5mm] at (anchor) {$\uparrow$};
      }{}
      \ifthenelse{\isin{#1}{right} \AND \isin{#2}{up}}{
            \coordinate (anchor) at ($({#3})!{#5}!({#4})$);
            \node[reddarkmsgcircle, xshift=5.0mm] at (anchor) {#6};
            \node[xshift=1.5mm] at (anchor) {$\uparrow$};
      }{}
}

\newcommand{\redmsg}[6]{
      \ifthenelse{\isin{#1}{left} \AND \isin{#2}{down}}{
            \coordinate (anchor) at ($({#3})!{#5}!({#4})$);
            \node[redmsgcircle, xshift=-5mm] at (anchor) {#6};
            \node[xshift=-1.5mm] at (anchor) {$\downarrow$};
      }{}
      \ifthenelse{\isin{#1}{right} \AND \isin{#2}{down}}{
            \coordinate (anchor) at ($({#3})!{#5}!({#4})$);
            \node[redmsgcircle, xshift=5mm] at (anchor) {#6};
            \node[xshift=1.5mm] at (anchor) {$\downarrow$};
      }{}

      \ifthenelse{\isin{#1}{down} \AND \isin{#2}{right}}{
            \coordinate (anchor) at ($({#3})!{#5}!({#4})$);
            \node[redmsgcircle, yshift=-5.0mm] at (anchor) {#6};
            \node[yshift=-2.0mm] at (anchor) {$\rightarrow$};
      }{}
      \ifthenelse{\isin{#1}{up} \AND \isin{#2}{right}}{
            \coordinate (anchor) at ($({#3})!{#5}!({#4})$);
            \node[redmsgcircle, yshift=5.0mm] at (anchor) {#6};
            \node[yshift=2.0mm] at (anchor) {$\rightarrow$};
      }{}

      \ifthenelse{\isin{#1}{down} \AND \isin{#2}{left}}{
            \coordinate (anchor) at ($({#3})!{#5}!({#4})$);
            \node[redmsgcircle, yshift=-5.0mm] at (anchor) {#6};
            \node[yshift=-2.0mm] at (anchor) {$\leftarrow$};
      }{}
      \ifthenelse{\isin{#1}{up} \AND \isin{#2}{left}}{
            \coordinate (anchor) at ($({#3})!{#5}!({#4})$);
            \node[redmsgcircle, yshift=5.0mm] at (anchor) {#6};
            \node[yshift=2.0mm] at (anchor) {$\leftarrow$};
      }{}

      \ifthenelse{\isin{#1}{left} \AND \isin{#2}{up}}{
            \coordinate (anchor) at ($({#3})!{#5}!({#4})$);
            \node[redmsgcircle, xshift=-5.0mm] at (anchor) {#6};
            \node[xshift=-1.5mm] at (anchor) {$\uparrow$};
      }{}
      \ifthenelse{\isin{#1}{right} \AND \isin{#2}{up}}{
            \coordinate (anchor) at ($({#3})!{#5}!({#4})$);
            \node[redmsgcircle, xshift=5.0mm] at (anchor) {#6};
            \node[xshift=1.5mm] at (anchor) {$\uparrow$};
      }{}
}

\newcommand{\bwmsg}[6]{
      \ifthenelse{\isin{#1}{left} \AND \isin{#2}{down}}{
            \coordinate (anchor) at ($({#3})!{#5}!({#4})$);
            \node[msgdoublecircle, xshift=-5.5mm] at (anchor) {#6};
            \node[xshift=-1.5mm] at (anchor) {$\downarrow$};
      }{}
      \ifthenelse{\isin{#1}{right} \AND \isin{#2}{down}}{
            \coordinate (anchor) at ($({#3})!{#5}!({#4})$);
            \node[msgdoublecircle, xshift=5.5mm] at (anchor) {#6};
            \node[xshift=1.5mm] at (anchor) {$\downarrow$};
      }{}

      \ifthenelse{\isin{#1}{down} \AND \isin{#2}{right}}{
            \coordinate (anchor) at ($({#3})!{#5}!({#4})$);
            \node[msgdoublecircle, yshift=-6.0mm] at (anchor) {#6};
            \node[yshift=-2.0mm] at (anchor) {$\rightarrow$};
      }{}
      \ifthenelse{\isin{#1}{up} \AND \isin{#2}{right}}{
            \coordinate (anchor) at ($({#3})!{#5}!({#4})$);
            \node[msgdoublecircle, yshift=6.0mm] at (anchor) {#6};
            \node[yshift=2.0mm] at (anchor) {$\rightarrow$};
      }{}

      \ifthenelse{\isin{#1}{down} \AND \isin{#2}{left}}{
            \coordinate (anchor) at ($({#3})!{#5}!({#4})$);
            \node[msgdoublecircle, yshift=-6.0mm] at (anchor) {#6};
            \node[yshift=-2.0mm] at (anchor) {$\leftarrow$};
      }{}
      \ifthenelse{\isin{#1}{up} \AND \isin{#2}{left}}{
            \coordinate (anchor) at ($({#3})!{#5}!({#4})$);
            \node[msgdoublecircle, yshift=6.0mm] at (anchor) {#6};
            \node[yshift=2.0mm] at (anchor) {$\leftarrow$};
      }{}

      \ifthenelse{\isin{#1}{left} \AND \isin{#2}{up}}{
            \coordinate (anchor) at ($({#3})!{#5}!({#4})$);
            \node[msgdoublecircle, xshift=-5.5mm] at (anchor) {#6};
            \node[xshift=-1.5mm] at (anchor) {$\uparrow$};
      }{}
      \ifthenelse{\isin{#1}{right} \AND \isin{#2}{up}}{
            \coordinate (anchor) at ($({#3})!{#5}!({#4})$);
            \node[msgdoublecircle, xshift=5.5mm] at (anchor) {#6};
            \node[xshift=1.5mm] at (anchor) {$\uparrow$};
      }{}
}

\newcommand{\bwdarkmsg}[6]{
      \ifthenelse{\isin{#1}{left} \AND \isin{#2}{down}}{
            \coordinate (anchor) at ($({#3})!{#5}!({#4})$);
            \node[darkmsgdoublecircle, xshift=-5.5mm] at (anchor) {#6};
            \node[xshift=-1.5mm] at (anchor) {$\downarrow$};
      }{}
      \ifthenelse{\isin{#1}{right} \AND \isin{#2}{down}}{
            \coordinate (anchor) at ($({#3})!{#5}!({#4})$);
            \node[darkmsgdoublecircle, xshift=5.5mm] at (anchor) {#6};
            \node[xshift=1.5mm] at (anchor) {$\downarrow$};
      }{}

      \ifthenelse{\isin{#1}{down} \AND \isin{#2}{right}}{
            \coordinate (anchor) at ($({#3})!{#5}!({#4})$);
            \node[darkmsgdoublecircle, yshift=-6.0mm] at (anchor) {#6};
            \node[yshift=-2.0mm] at (anchor) {$\rightarrow$};
      }{}
      \ifthenelse{\isin{#1}{up} \AND \isin{#2}{right}}{
            \coordinate (anchor) at ($({#3})!{#5}!({#4})$);
            \node[darkmsgdoublecircle, yshift=6.0mm] at (anchor) {#6};
            \node[yshift=2.0mm] at (anchor) {$\rightarrow$};
      }{}

      \ifthenelse{\isin{#1}{down} \AND \isin{#2}{left}}{
            \coordinate (anchor) at ($({#3})!{#5}!({#4})$);
            \node[darkmsgdoublecircle, yshift=-6.0mm] at (anchor) {#6};
            \node[yshift=-2.0mm] at (anchor) {$\leftarrow$};
      }{}
      \ifthenelse{\isin{#1}{up} \AND \isin{#2}{left}}{
            \coordinate (anchor) at ($({#3})!{#5}!({#4})$);
            \node[darkmsgdoublecircle, yshift=6.0mm] at (anchor) {#6};
            \node[yshift=2.0mm] at (anchor) {$\leftarrow$};
      }{}

      \ifthenelse{\isin{#1}{left} \AND \isin{#2}{up}}{
            \coordinate (anchor) at ($({#3})!{#5}!({#4})$);
            \node[darkmsgdoublecircle, xshift=-5.5mm] at (anchor) {#6};
            \node[xshift=-1.5mm] at (anchor) {$\uparrow$};
      }{}
      \ifthenelse{\isin{#1}{right} \AND \isin{#2}{up}}{
            \coordinate (anchor) at ($({#3})!{#5}!({#4})$);
            \node[darkmsgdoublecircle, xshift=5.5mm] at (anchor) {#6};
            \node[xshift=1.5mm] at (anchor) {$\uparrow$};
      }{}
}

\tikzset{mainstyle/.style={fill=white, draw=black, shape=rectangle, align=center}}

\tikzset{dstyle/.style={mainstyle, minimum size=4mm, inner sep=0pt, text width=4mm}}

\tikzset{sstyle/.style={mainstyle, minimum size=5mm, inner sep=0pt, text width=5mm}}

\tikzset{ostyle/.style={fill=darkgrey, draw=black, shape=rectangle, minimum size=0.2cm, inner sep=0pt, text width=2mm}}

\tikzstyle{observation}=[ostyle]
\tikzstyle{deterministic}=[dstyle]
\tikzstyle{stochastic}=[sstyle]

\tikzstyle{filter}=[mainstyle, minimum width=1cm, minimum height=0.5cm]
\tikzstyle{selector}=[fill=white, draw=black, shape=trapezium, rotate=180, minimum width=1cm, minimum height=0.5cm]

\newcommand*\wcircled[1]{\tikz[baseline=(char.base)]{
            \node[shape=circle,draw,minimum size=4mm,inner sep=0pt] (char) {#1};}}

\def\e{{\mathbf e}}
\def\u{{\mathbf u}}

\def\y{{\mathbf y}}

\def\d{{\mathrm d}}
\def\-{\text{-}}
\def\+{\text{+}}
\newcommand\given[1][]{\:#1\vert\:}

\title{\LARGE \bf
Variational message passing for\\ online polynomial NARMAX identification
}

\author{Wouter M. Kouw, Albert Podusenko, Magnus T. Koudahl and Maarten Schoukens%
\thanks{Kouw, Podusenko and Koudahl are with the Bayesian Intelligent Autonomous Systems lab and Schoukens is with the Control Systems group, all are part of the department of Electrical Engineering, TU Eindhoven, Eindhoven, the Netherlands. {\tt\small @: w.m.kouw@tue.nl}}%
}

\begin{document}

\maketitle
\thispagestyle{empty}
\pagestyle{empty}

\begin{abstract}
We propose a variational Bayesian inference procedure for online nonlinear system identification. For each output observation, a set of parameter posterior distributions is updated, which is then used to form a posterior predictive distribution for future outputs. We focus on the class of polynomial NARMAX models, which we cast into probabilistic form and represent in terms of a Forney-style factor graph. Inference in this graph is efficiently performed by a variational message passing algorithm. We show empirically that our variational Bayesian estimator outperforms an online recursive least-squares estimator, most notably in small sample size settings and low noise regimes, and performs on par with an iterative least-squares estimator trained offline.
\end{abstract}
\section{Introduction}
Nonlinear autoregressive moving-average with exogenous input (NARMAX) models are a staple in modern system identification. They have been applied to a wide range of systems such as the kinematics of mobile robots, the effects of space weather on earthbound electronics or the visual system of fruit flies \cite{billings2013nonlinear}. We are interested in online estimators, because they allow for in-situ learning on embedded systems, and Bayesian estimators, i.e., posterior distributions instead of point estimates \cite{murphy2012machine,sarkka2013bayesian}. The advantage of Bayesian estimators is that they are naturally robust to overfitting when data is still scarce \cite[Ch.~5.3.1]{murphy2012machine}. This paper proposes a recursive approximate Bayesian estimator for online system identification.

Despite its long history, Bayesian identification has always been challenging from a practical perspective \cite{peterka1981bayesian}. Intractable integrals may prevent the formulation of an exact Bayesian estimator. 
Approximate Bayesian inference, especially Sequential Monte Carlo (a.k.a. particle filtering), has proven to be much more practical for dynamical systems \cite{schon2015sequential}. Nonetheless, Monte Carlo-based methods are still quite computationally expensive.
Variational Bayesian inference is an attractive alternative because it is typically much cheaper - computation-wise - than Monte Carlo sampling. The unnormalized posterior distribution function is approximated by minimizing a variational free energy functional with respect to a second probabilistic model \cite{blei2017variational}. The first uses of variational Bayes for system identification allowed for simultaneous estimation of states, coefficients and noise parameters in a wide range nonlinear stochastic differential equations \cite{friston2008variational,daunizeau2009variational}. A particular technique called Dynamic Expectation Maximization (DEM), became popular and was recently used to simultaneously estimate not only states and inputs, but also colored noise \cite{friston2008variational,meera2020free}. DEM relies on Laplace's method, i.e., approximating the posterior with a Gaussian distribution using gradient-based techniques for finding the mode and local curvature. However, Laplace approximations fail for non-modal, multi-modal or discrete distributions, and can be inaccurate for distributions with higher-order moments (e.g., skewed or kurtotic ones). We employ a richer class of form constraints on the approximating distributions, namely the exponential family.
A few recent papers have ventured into non-parametric families such as Gaussian processes and deep neural networks, achieving impressive results \cite{risuleo2017variational,hendriks2021deep}. But non-parametric models can quickly become computationally costly again. We formulate the inference procedure as message passing on a factor graph \cite{loeliger2004introduction}. Computation can be distributed along nodes and edges by exploiting the factorization of the probabilistic model. This produces an efficient and parallelizable algorithm \cite{korl2005factor,dauwels2007variational}.
Lastly, variational Bayes has found its way to autoregressive-based models. A recent ARMAX paper infers the noise sequence explicitly, but extending it to the nonlinear case is not trivial \cite{fujimoto2016system}. In addition, there are also NARX, NLARX, and SARX models which show competitive performance \cite{lu2014variational,kouw2020online,jacobs2018sparse}. Our work complements these techniques by extending the scope to polynomial NARMAX models. 
%

Our key contribution is the formulation of a recursive parameter and posterior predictive estimation algorithm using variational message passing on a Forney-style factor graph (Sec.~\ref{sec:factor-graph}). We show that our estimator competes well with an online least-squares estimator, outperforming it in small sample size settings without the need for informative priors (Sec.~\ref{sec:experiments}). 

\section{NARMAX system}
Consider a discrete-time dynamical system with an unknown time horizon, indexed by time $ \in \mathbb{N}$. Let $u_k \in \mathbb{R}$ be a measured input signal, $y_k \in \mathbb{R}$ a measured output signal and $e_k \in \mathbb{R}$ be noise, drawn from a zero-mean Gaussian distribution with zero auto-correlation: $e_k \sim \mathcal{N}(0, \tau^{-1})$ where $\tau$ is a precision (inverse variance) parameter. In a NARMAX system, the output $y_k$ is generated according to:
\begin{align}
    y_k = f(u_k, \u_{k-1}, \y_{k-1}, \e_{k-1}) + e_k \, ,
\end{align}
where $\u_{k-1} = (u_{k-1}, ..\, u_{k-M_1})$ is a vector containing $M_1$ delayed inputs, $\y_{k-1} = (y_{k-1}, ..\, y_{k-M_2})$ contains $M_2$ delayed outputs and the vector $\e_{k-1} = (e_{k-1 }, ..\, e_{k-M_3})$ contains $M_3$ delayed noise instances. 
The function $f$ is assumed to be continuous, nonlinear, and time-invariant.

\section{Probabilistic Model} \label{sec:model}
In a polynomial NARMAX, the function $f$ is modeled with a linear combination of coefficients $\theta$ and a polynomial basis function $\phi$ applied to inputs, outputs and errors:
\begin{align} \label{eq:polnarmax}
    y_k = \theta^{\top} \phi(u_k, \u_{k-1}, \y_{k-1}, \e_{k-1}) + e_k \, .
\end{align}
We define the vector $\phi_k = \phi(u_k, \u_{k-1}, \y_{k-1}, \e_{k-1})$ for conciseness in later derivations. Specifying a probabilistic model consists of expressing the likelihood of observations, given parameters and noise, and posing a set of prior distributions for the unknown variables. 

\subsection{Likelihood function} \label{sec:likelihood}
The noise variable is Gaussian distributed, which lets us express the likelihood of observing $y_k$ as:
\begin{align}
    p(y_k \given u_k, \u_{k\-1}, \y_{k\-1}, \e_{k\-1}, \theta, \tau)
    = \mathcal{N}\big(y_k \given \theta^{\top} \phi_k, \tau^{-1}\big) . \label{eq:likelihood}
\end{align}
In this notation, it is implied that variables with subscripts smaller than $1$ drop out. So, the likelihood of the first observation simplifies to $p(y_1 \given u_1, \theta, \tau)$. In practice, the vectors $\u_{k-1}$, $\y_{k-1}$, and $\e_{k-1}$ can be initialized with zeros and updated as data streams in. This allows for the recursive application of \eqref{eq:likelihood}. 

\subsection{Prior distributions} \label{sec:priors}
Our model has two unknown variables: the coefficients $\theta$ and the noise precision $\tau$. We need to pose an initial prior distribution for each. The coefficients are unbounded real-valued numbers, which could be modeled with a variety of continuous probability distributions. We choose a Gaussian distribution because its linear transformation $\theta^{\top}\phi_k$ results in another Gaussian that is conditionally conjugate to the likelihood function in \eqref{eq:likelihood} \cite[Ch.~4.6]{murphy2012machine}. 
%
The precision parameter $\tau$ is a strictly positive number, which could be modeled with for instance an Exponential or Gamma distribution. We choose a Gamma distribution, also because it is conditionally conjugate to our Gaussian likelihood \cite[Ch.~4.6]{murphy2012machine}.
%
The initial priors are denoted as:
\begin{align} \label{eq:priors}
    p(\theta) = \mathcal{N}\big(\theta \given \mu_0, \Lambda_0^{-1}\big) \, , \quad \
    p(\tau) = {\it \Gamma}\big(\tau \given \alpha_0, \beta_0 \big) \, ,
\end{align}
where the subscripts refer to time $k=0$, i.e., before $k=1$.
We parameterize our Gaussian distributions with means $\mu$ and precision matrices $\Lambda$ (inverse covariance matrix) and our Gamma distributions with shapes $\alpha$ and rates $\beta$.

\subsection{Parameter posteriors} \label{sec:param-post}
Given a likelihood function and prior distributions, we can apply Bayes' rule to obtain posterior distributions. For the purposes of online system identification, we describe the posterior recursively \cite[Chapter~3]{sarkka2013bayesian}. We start with the initial application of Bayes' rule:
\begin{align}
    \underbrace{p(\theta, \tau \given y_1, u_1)}_{\text{posterior at $k$=$1$}} = \underbrace{\frac{1}{p(y_1 \given u_1)}}_{\text{evidence}} \underbrace{p(y_1 \given u_1, \theta, \tau)}_{\text{likelihood}} \underbrace{p(\theta) p( \tau)}_{\text{initial priors}} .
\end{align}
The likelihood is multiplied with both priors to form a joint distribution over $y_1$, $\theta$ and $\tau$. That joint is normalized by the evidence for $y_1$, after which a joint posterior distribution for the parameters is obtained. 

In recursive estimation, the posterior at one time point becomes the prior for the next \cite{sarkka2013bayesian}. At $k=2$, we have:
\begin{align}
     &\underbrace{p(\theta, \tau  \given y_{1:2}, u_{1:2}, e_1)}_{\text{posterior at $k$=$2$}} = \underbrace{\frac{1}{p(y_2 \given u_{1:2}, y_1, e_1)}}_{\text{evidence for $y_2$}} \nonumber \\
    &\ \qquad \qquad \cdot \underbrace{p(y_2 \given u_{1:2}, y_{1}, e_1, \theta, \tau)}_{\text{likelihood of $y_2$}} \underbrace{p(\theta, \tau \given y_1, u_1)}_{\text{prior (posterior $k$=$1$)}} \, .
\end{align}
The likelihood now contains the first elements of the previous input $\u_{k-1}$, output $\y_{k-1}$ and error $\e_{k-1}$ vectors. 
Note the structure of this equation: the previous posterior distribution is updated using two terms describing properties of the new observation $y_2$. 
In general, at time $k$, we have the following recursive posterior estimation procedure:
\begin{align} \label{eq:recursive}
    & \underbrace{p(\theta, \tau | y_{1:k}, u_{1:k}, e_{1:k\-1})}_{\text{parameter posterior at $k$}} = \frac{1}{\underbrace{p(y_k | u_{1:k}, y_{1:k\-1}, e_{1:k\-1})}_{\text{evidence}}} \nonumber \\
     & \cdot \underbrace{p(y_k | u_k, \u_{k\-1}, \y_{k\-1}, \e_{k\-1}, \theta, \tau)}_{\text{NARMAX likelihood}} \underbrace{p(\theta, \tau | y_{1:k\-1}, u_{1:k\-1}, e_{1:k\-2})}_{\text{prior (posterior at $k\-1$)}} \, , 
\end{align}
where the evidence consists of integrating the product of the NARMAX likelihood and the prior with respect to the parameters $\theta$ and $\tau$:
\begin{align}
    p(y_k \given u_{1:k}, y_{1:k\-1}, &e_{1:k\-1}) = \nonumber \\
    \iint p(y_k \given u_k, &\u_{k\-1}, \y_{k\-1}, \e_{k\-1}, \theta, \tau) \nonumber \\
    & \cdot p(\theta, \tau \given y_{1:k\-1}, u_{1:k\-1}, e_{1:k\-2}) \ \d \theta\d \tau \, .
\end{align}
Unfortunately, the resulting posterior distribution is not of exactly the same form as the prior and is therefore not suited to recursive estimation. We approximate it with a more suitable distribution in Section \ref{sec:inference}.

\subsection{Posterior predictive} \label{sec:post-pred}
Given a posterior distribution of the parameters, the one-step-ahead posterior predictive distribution for the output is:
\begin{align} \label{eq:post-pred}
    &\underbrace{p(y_{k+1} \given u_{1:k+1}, y_{1:k}, e_{1:k})}_{\text{posterior predictive}} = \nonumber \\
    &\qquad \iint \underbrace{p(y_{k+1} \given u_{k+1}, \u_k, \y_k, \e_k, \theta, \tau)}_{\text{likelihood of future observation}} \nonumber \\
    &\qquad \qquad \qquad \qquad \quad \cdot \underbrace{p(\theta, \tau \given y_{1:k}, u_{1:k}, e_{1:k\-1})}_{\text{parameter posterior}} \mathrm{d}\theta \mathrm{d}\tau \, . 
\end{align}
The posterior predictive is the average distribution for $y_{k+1}$, weighted by the posterior probability of each value of the parameters. This weighted average has a greater uncertainty than what would have obtained by plugging in a selected parameter. As such, the posterior predictive distribution is naturally regularized and is more robust to overfitting on the training data. Details on how to compute the posterior predictive are described in Section \ref{sec:forecasting}.

\subsection{Prediction errors} \label{sec:errors}
Typically, the prediction errors are defined as the difference between the observed output $y_{k+1}$ and a numerical prediction $\hat{y}_{k+1}$ based on previous data \cite{billings2013nonlinear}:
\begin{align}
    e_{k+1} = y_{k+1} - \hat{y}_{k+1} \, .
\end{align}
However, our prediction comes in the form of a posterior predictive distribution (i.e., a random variable, not a number). 
To adhere to the original definition of the prediction errors, we select the maximum a posteriori (MAP) of the posterior predictive distribution:
\begin{align}
    \hat{y}_{k+1} = \underset{y_{k+1}}{\arg \max} \ p(y_{k+1} \given u_{1:k+1}, y_{1:k}, e_{1:k}) \, .
\end{align}
To be clear, the order of operations in our recursive estimation procedure is as follows: at time $k$, we observe $y_k$ and update the parameter posterior according to \eqref{eq:recursive}. We then use the prediction $\hat{y}_k$ made during the previous time-step to compute the prediction error $e_k$. This error is used when we make a prediction for $\hat{y}_{k+1}$, which is passed on to the next time-step.


\section{Inference} \label{sec:inference}
It is not possible to obtain the posterior distribution exactly due to the priors being merely conditionally conjugate and not jointly conjugate to our NARMAX likelihood. Below, we show how to approximate it in a recursive manner.

\subsection{Free energy minimization} \label{sec:fem}
We adhere to a form of approximate Bayesian inference called variational free energy minimization \cite{blei2017variational}. 
Essentially, one poses a second probabilistic model $q$, called the \emph{recognition} model, with which the \emph{generative} model $p$ is approximated. The free energy functional at time $k$ is the Kullback-Leibler (KL) divergence between the recognition model and the true posterior, minus the log evidence:
\begin{align} \label{eq:fe_decomp1}
    \mathcal{F}_k[q_k] =& \underbrace{\iint q_k(\theta, \tau) \ln \frac{q_k(\theta, \tau)}{p(\theta, \tau \given y_{1:k}, u_{1:k}, e_{1:k\-1})} \d \theta \d \tau}_{\text{approximation of posterior}} \nonumber \\
    &\ \qquad - \ln \underbrace{p(y_k \given u_{1:k}, y_{1:k-1}, e_{1:k\-1})}_{\text{evidence}} \, .
\end{align} 
Note that the $q_k$ that minimizes $\mathcal{F}_k$ is an optimal approximation of the true posterior at time $k$. 

Equation \eqref{eq:fe_decomp1} necessitates the computation of the true posterior, which is intractable. We therefore re-formulate the objective along the lines of \eqref{eq:recursive}:
\begin{align} \label{eq:fe_decomp2}
    & \mathcal{F}_k[q_k] = \! \underbrace{\iint q_k(\theta, \tau) \ln \frac{q_k(\theta, \tau)}{p(\theta, \tau | y_{1:k\-1}, u_{1:k\-1}, e_{1:k\-2})} \d \theta \d \tau}_{\text{complexity}} \nonumber \\
    & \! - \! \underbrace{\iint \!q_k(\theta, \tau) \ln p(y_k | u_k, \u_{k\-1}, \y_{k\-1}, \e_{k\-1}, \theta, \tau) \d \theta \d \tau}_{\text{accuracy}} .
\end{align} 
Accuracy expresses how well the observation was predicted given the current parameter estimates and complexity is a measure of how much the recognition model deviates from the previous posterior. Minimizing $\mathcal{F}_k$ should therefore be interpreted as balancing a fit to data and avoiding large changes to parameters.

\subsection{Mean field assumption}
If we make a mean-field assumption on the factorization of the recognition model:
\begin{align} \label{eq:mean-field}
    q_k(\theta, \tau) = q_k(\theta) \, q_k(\tau) \, .
\end{align}
then we can derive the forms of the recognition factors for which $\mathcal{F}_k$ is minimal \cite{blei2017variational,dauwels2007variational}:
\begin{subequations} \label{eq:recursive-q}
\begin{align} 
    q_k&(\theta) \propto \underbrace{\exp \big(\mathbb{E}_{q_k(\tau)} \ln p(\theta, \tau \given y_{1:k\-1}, u_{1:k\-1}, e_{1:k\-2}) \big)}_{\wcircled{1}\ \text{prior-based}} \nonumber \\
    & \cdot \underbrace{\exp \big(\mathbb{E}_{q_k(\tau)} \ln p(y_k \given u_k, \u_{k\-1}, \y_{k\-1}, \e_{k\-1}, \theta, \tau) \big)}_{\wcircled{2}\ \text{likelihood-based}} \, , \label{eq:recursive-q-theta}\\
    q_k&(\tau) \propto \underbrace{\exp \big(\mathbb{E}_{q_k(\theta)} \ln p(\theta, \tau \given y_{1:k\-1}, u_{1:k\-1}, e_{1:k\-2}) \big] \big)}_{\wcircled{3}\ \text{prior-based}} \nonumber \\
    & \cdot \underbrace{\exp \big(\mathbb{E}_{q_k(\theta)} \ln p(y_k \given u_k, \u_{k\-1}, \y_{k\-1}, \e_{k\-1}, \theta, \tau) \big)}_{\wcircled{4}\ \text{likelihood-based}} . \label{eq:recursive-q-tau}
\end{align}
\end{subequations}
At $k$ = $0$, the prior-based terms $\wcircled{1}$ and $\wcircled{3}$ correspond directly to the prior distributions in \eqref{eq:priors}. 
%
Computing and updating these recognition factors can be formulated as a variational message passing algorithm \cite{dauwels2007variational}.

\subsection{Message passing on factor graphs} \label{sec:factor-graph}
\begin{figure}[htb]
    \centering
    \scalebox{1.2}{\begin{tikzpicture}

    \node [style=deterministic] (phi) {$\phi$};
    \node [style=deterministic, right=5mm of phi] (dot) {$\cdot$};
    \node [style=stochastic, right=10mm of dot] (N) {$\mathcal{N}$};

    \node [style=observation, below left = 1mm and 8mm of phi] (us) {};
    \node [left of=us, node distance=5mm] (uss) {$\scriptstyle{\mathbf{u}_{k-1}}$};
    \node [style=observation, above left = 1mm and 8mm of phi] (es) {};
    \node [left of=es, node distance=5mm] (ess) {$\scriptstyle{\mathbf{e}_{k-1}}$};
    \node [style=observation, left = 8mm of phi] (ys) {};
    \node [left of=ys, node distance=5mm] (yss) {$\scriptstyle{\mathbf{y}_{k-1}}$};
    \node [style=observation, below = 7mm of phi] (u) {};
    \node [below of=u, node distance=4mm] (uk) {$\scriptstyle{u_k}$};

    \node [style=observation, below = 7mm of N] (y) {};
    \node [below of=y, node distance=4mm] (yk) {$\scriptstyle{y_k}$};

    \node [style=deterministic, above = 12mm of dot] (eqN) {$=$};
    \node [left = 18mm of eqN] (prevtheta) {$\cdots$};

    \node [style=deterministic, above = 20mm of N] (eqG) {$=$};
    \node [left = 33mm of eqG] (prevtau) {$\cdots$};

    \node [style=stochastic, left = 5mm of prevtheta] (thetaprior) {$\mathcal{N}$};
    \node [style=stochastic, left = 5mm of prevtau] (tauprior) {$\Gamma$};

    \node [right = 22mm of eqN] (thetapost) {$\cdots$};
    \node [right = 7mm of eqG] (taupost) {$\cdots$};
    
    \node[dashed, fit=(phi)(N), draw, inner sep=4mm] (box) {};

    \draw [->] (phi) -- (dot);
    \draw [->] (dot) -- (N);
    \draw [->] (eqG) -- (N);
    \draw [->] (eqN) -- (dot);
    \draw [->] (N) -- (y);
    \draw [->] (u) -- (phi);

    \draw [-] (tauprior) -- (prevtau);
    \draw [-] (prevtau) -- (eqG);
    \draw [-] (thetaprior) -- (prevtheta);
    \draw [-] (prevtheta) -- (eqN);
    \draw [-] (eqN) -- (thetapost);
    \draw [-] (eqG) -- (taupost);
    \node [above of=eqN, node distance=4mm] {$\scriptstyle{\theta}$};
    \node [above of=eqG, node distance=4mm] {$\scriptstyle{\tau}$};

    \draw[->] (us.east) -| ($(phi.south)$);
    \draw[->] (ys.east) -- ($(phi.west)$);
    \draw[->] (es.east) -| ($(phi.north)$);

    \msgcircle{up}{right}{prevtheta}{eqN}{0.5}{1};
    \msgcircle{up}{right}{prevtau}{eqG}{0.6}{3};
    \msgcircle{left}{up}{dot}{eqN}{0.6}{2};
    \msgcircle{left}{up}{N}{eqG}{0.4}{4};



\end{tikzpicture}}
    \caption{Forney-style factor graph of the polynomial NARMAX model in recursive form. Prior-based messages $1$ and $3$ enter from the left (previous time-step). Likelihood-based messages $2$ and $4$ are passed upwards from the composite "NARMAX" node (dotted box), which is attached to observed variables $y_k$, $u_k$, $\u_{k\-1}$, $\y_{k\-1}$ and $\e_{k\-1}$. At the equality nodes, the recognition factors are updated based on combining the prior-based and likelihood-based messages. }
    \label{fig:ffg}
\end{figure}
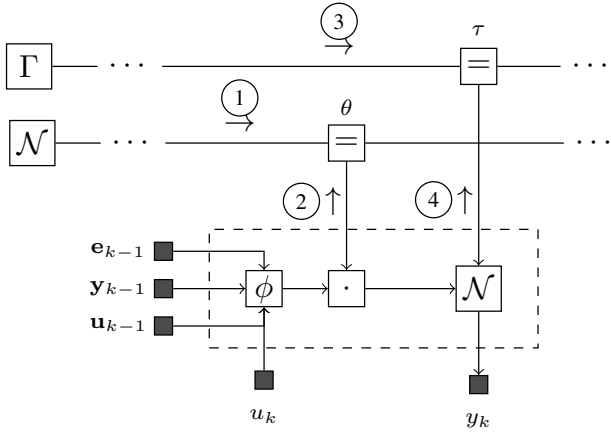

Factor graphs are visual representations of probabilistic models \cite{loeliger2004introduction}. Figure \ref{fig:ffg} shows a Forney-style factor graph (FFG) of the probabilistic NARMAX model in recursive form. The square nodes represent operations, either deterministic such as the basis expansion or the dot product, or stochastic such as the Gaussian and Gamma prior distributions. Edges represent unknown variables with associated recognition factors, except those terminated by small black squares as they correspond to observed variables. Nodes containing an "=" sign represent an equality constraint posed on all connected edges \cite{loeliger2004introduction}. The dotted box is a composite node encompassing all the operations in the NARMAX likelihood.

The inference procedure starts with the nodes on the left (initial priors) which pass messages rightwards towards the two equality nodes. Each time-step, the messages containing prior information, $\wcircled{1}$ and $\wcircled{3}$, travel downwards from the equality node and arrive at the composite NARMAX likelihood node. The composite node first incorporates all observed variables and performs its internal operations. Then, it uses incoming message $\wcircled{1}$ to pass message $\wcircled{4}$ along the edge corresponding to the noise precision variable. It also uses message $\wcircled{3}$ to pass message $\wcircled{2}$ towards the coefficients. 

The equality nodes perform the recognition factor updates: the prior-based messages from the left, $\wcircled{1}$ and $\wcircled{3}$, and likelihood-based messages from below, $\wcircled{2}$ and $\wcircled{4}$, are combined according to \eqref{eq:recursive-q}. These updated beliefs are then passed downwards again, where the NARMAX node uses them to compute new outgoing messages. After a prespecified number of iterations, message passing is halted and the resulting recognition factors are sent rightwards to serve as priors for the next time-step.

\subsection{Variational messages}
We impose the constraint that each recognition factor belongs to a parametric family of distributions. For ease of computation, we choose the following families:
\begin{align} \label{eq:recognition}
    q_k(\theta) = \mathcal{N}\big(\theta \given \mu_k, \Lambda_k^{-1} \big) \, , \quad q_k(\tau) = {\it \Gamma} \big(\tau \given \alpha_k, \beta_k \big) \, .
\end{align}
This constraint alters $\mathcal{F}_k$ from a functional to a function: it is now minimized with respect to the parameters $\mu_k$, $\Lambda_k$, $\alpha_k$ and $\beta_k$ instead of a general probability distribution $q_k$.

In order to obtain Messages $\wcircled{2}$ and $\wcircled{4}$, we need to take expectations with respect to each recognition factor\footnote{Detailed derivations as well as code for experiments can be found at \url{https://github.com/biaslab/ACC2022-vmpNARMAX}.}. For the coefficient recognition factor, i.e., \eqref{eq:recursive-q-theta}, this is:
\begin{align}
    \wcircled{2} =&\ \exp \big(\mathbb{E}_{q_k(\tau)} \ln p(y_k | u_k, \u_{k\-1}, \y_{k\-1}, \e_{k\-1}, \theta, \tau) \big) \nonumber \\
    \propto&\ \exp\Big(- \frac{1}{2} \frac{\alpha_k}{\beta_k} \big(-2y_k\theta^{\top}\phi_k + \theta^{\top} \phi_k \phi_k^{\top} \theta \big) \Big) \, .
\end{align}
One may recognize a Gaussian probability density function $\mathcal{N}(\theta \given \bar{\mu}_k, \bar{\Lambda}^{-1}_k)$ with parameters:
\begin{align} \label{eq:update_theta}
    \bar{\mu}_k = \Big(\frac{\alpha_k}{\beta_k} \phi_k \phi_k^{\top} \Big)^{-1} \frac{\alpha_k}{\beta_k} y_k \phi_k \, , \quad 
     \bar{\Lambda}_{k} = \frac{\alpha_k}{\beta_k} \phi_k \phi_k^{\top} \, .
\end{align}
One may alternatively parameterize this Gaussian in terms of the precision and precision-weighted mean, as is done in information filters \cite{sarkka2013bayesian}. This avoids a matrix inversion during the recognition factor update (Sec.~\ref{sec:updating-marginals}).

The likelihood-based term for the precision in \eqref{eq:recursive-q-tau} is:
\begin{align}
    & \wcircled{4} = \exp \big(\mathbb{E}_{q_k(\theta)} \ln p(y_k | u_k, \u_{k\-1}, \y_{k\-1}, \e_{k\-1}, \theta, \tau) \big) \nonumber \\
    & \propto \tau^{1/2} \exp\Big(- \! \frac{\tau}{2} \big((y_k - \mu^{\top}\phi_k)^2 + \phi_k^{\top} \Lambda^{-1}_k \phi_k \big) \Big) \, .
\end{align}
One may recognize the probability density function of a Gamma distribution ${\it \Gamma}(\tau \given \bar{\alpha}_k, \bar{\beta}_k)$ with parameters:
\begin{align} \label{eq:update_tau}
    \bar{\alpha}_{k} \! = \! \frac{3}{2} \, , \quad
    \bar{\beta}_{k} \! = \! \frac{1}{2}\big((y_k - \mu^{\top}\phi_k)^2 + \phi_k^{\top} \Lambda^{-1}_k \phi_k \big) \, .
\end{align}

At $k=1$, messages $\wcircled{1}$ and $\wcircled{3}$ consist of the initial priors in \eqref{eq:priors}. Afterwards, the posteriors are approximated by recognition factors. That means the prior-based term in \eqref{eq:recursive-q-theta} simplifies to:
\begin{align}
    \wcircled{1} =&\ \exp \big(\mathbb{E}_{q_k(\tau)} \ln p(\theta, \tau \given y_{1:k-1}, u_{1:k-1}, e_{1:k-2}) \big) \nonumber \\
    \approx&\ \exp \big(\mathbb{E}_{q_k(\tau)} \ln \big[ q_{k\-1}(\theta) q_{k\-1}(\tau) \big] \big) 
    \propto q_{k\-1}(\theta) \, .
\end{align}
Similarly, the prior-based term in \eqref{eq:recursive-q-tau} simplifies to:
\begin{align}
    \wcircled{3} =&\ \exp \big(\mathbb{E}_{q_k(\theta)} \ln p(\theta, \tau \given y_{1:k-1}, u_{1:k-1}, e_{1:k-2}) \big) \nonumber \\
    \approx&\ \exp \big(\mathbb{E}_{q_k(\theta)} \ln \big[ q_{k\-1}(\theta) q_{k\-1}(\tau) \big] \big) 
    \propto q_{k\-1}(\tau) \, .
\end{align}

\subsection{Updating recognition factors} \label{sec:updating-marginals}
The combination of $\wcircled{1}$ and $\wcircled{2}$ is the product of two Gaussian probability density functions which is proportional to another Gaussian density $\mathcal{N}(\theta \given \mu_k, \Lambda_k^{-1})$ where:
\begin{flalign} \label{eq:params_theta}
    & \quad \Lambda_k \! = \! \Lambda_{k-1} \! + \!  \bar{\Lambda}_{k} \, , \quad \
    \Lambda_k \mu_k \! = \! \Lambda_{k-1} \mu_{k-1} \! + \! \bar{\Lambda}_{k} \bar{\mu}_{k}  \, .
\end{flalign}
Since $\bar{\Lambda}_k$ is strictly positive, the precision of the recognition factor always grows after making a new observation.

The combination of $\wcircled{3}$ and $\wcircled{4}$ is the product of two Gamma probability density functions and is proportional to another Gamma density ${\it \Gamma}(\tau \given \alpha_k, \beta_k)$ where:
\begin{align} \label{eq:params_tau}
    \alpha_k = \alpha_{k-1} + \bar{\alpha}_k - 1  \, , \qquad
    \beta_k = \beta_{k-1} + \bar{\beta}_k \, . 
\end{align}
The shape parameter grows by $1/2$ each time-step, since $\bar{\alpha}_k$ is always $3/2$. Although the rate parameter also always grows with more observations ($\bar{\beta}_k$ consists only of quadratic terms), the mean of $\tau$ can still shrink when $\beta_k$ grows at a slower pace than $\alpha_k$.

Equations \eqref{eq:recursive-q} describe optimal forms for the recognition factors, but these forms depend on each other: the updates to $\mu_k$ and $\Lambda_k$ depend on $\alpha_k$ and $\beta_k$ (\ref{eq:params_theta} and \ref{eq:recursive-q-theta}) and the update to $\beta_k$ depends on $\mu_k$ and $\Lambda_k$ (\ref{eq:params_tau} and \ref{eq:recursive-q-tau}). They must therefore be iterated until convergence. This form of variational inference is equivalent to an exact coordinate descent procedure: each recognition factor update is an exact minimization step with respect to the current variational parameters \cite{blei2017variational,dauwels2007variational}. The algorithm is guaranteed to converge because each update leads to an equal or smaller value of the free energy objective function \eqref{eq:fe_decomp2} \cite{yedidia2005constructing}.

\section{Model simulation} \label{sec:forecasting}
We compute the one-step ahead prediction from \eqref{eq:post-pred} using the approximate posteriors $q_k(\theta)$ and $q_k(\tau)$. At time $k$, the posterior predictive for $k+1$ is approximately:
\begin{align}
    p(y_{k+1} \given& u_{1:k+1}, y_{1:k}, e_{1:k})
    \nonumber \\
    &\approx \mathbb{E}_{q_k(\theta)} \mathbb{E}_{q_k(\tau)} \big[\,  \mathcal{N}(y_{k+1} \given \theta^{\top} \phi_{k+1}, \tau^{-1}) \big] \, .
\end{align}
The vector $\phi_{k+1}$ contains the next input $u_{k+1}$ and the vectors $\u_k$, $\y_k$ and $\e_k$. The expectation with respect to the precision parameter produces a Student's t-distribution with $2\alpha_k$ degrees of freedom \cite{murphy2012machine}. For computational convenience, we approximate this distribution with a Gaussian distribution with the same parameters:
\begin{align}
    \mathbb{E}_{q_k(\! \tau \!)} \! \big[\mathcal{N}(y_{k\+1} | \, \theta^{\top} \phi_{k\+1}, \tau^{\-1}) \big] \! \approx \!  \mathcal{N}(y_{k\+1} | \, \theta^{\top} \phi_{k\+1}, \frac{\beta_k}{\alpha_k}) .
\end{align}
Note that this approximation becomes tighter as $\alpha_k$ grows. The remaining expectation with respect to the coefficients is:
\begin{align} \label{eq:predictive}
    \mathbb{E}_{q_k(\theta)}&\big[\, \mathcal{N}(y_{k+1} \given \theta^{\top} \phi_{k+1}, \frac{\beta_k}{\alpha_k}) \big] \nonumber \\
    =&\ \mathcal{N} \big(y_{k+1} \given \mu_k^{\top} \phi_{k+1},\ \phi_{k+1}^{\top} \Lambda^{-1}_k \phi_{k+1} + \frac{\beta_k}{\alpha_k} \big) \, .
\end{align}

Simulations can be generated by fixing the parameters $\mu_k$, $\Lambda_k$, $\alpha_k$ and $\beta_k$ to their final estimates and then applying the mean and variance calculation from \eqref{eq:predictive} to $\phi_{i}$ for $i=1, \dots T$ time steps. Instead of observed output, the vector $\y_{i}$ will contain the MAP estimates of the posterior predictive distribution $\hat{y}_{i}$, produced during $i-1$. Instead of the prediction errors, the vector $\e_{i}$ will contain zeros. This zero-padding is a common technique for simulation with NARMAX models, but comes at the cost of a bias \cite{khandelwal2018simulation}. 

\section{Experiments} \label{sec:experiments}
We performed two experiments on data generated from a simulated NARMAX system: 1) the noise level is fixed while the length of the signal for training is varied, and 2) the training signal length is fixed while the noise level is varied.
Our Variational Message Passing (VMP) estimator was compared to two baselines: a Recursive Least-Squares (RLS) estimator with a forgetting factor of $1.0$ \cite[Sec.~9.4]{hayes2009statistical} and a Iterative Least-Squares (ILS) estimator trained offline \cite[Section~3.6]{billings2013nonlinear}. Since these lack posterior predictive distributions, we evaluate in terms of Root Mean Square (RMS) errors over a validation signal of length 1000.

\subsection{Data generation}
We generated a random-phase multisine input signal consisting of a range of 100 frequencies between 0 to 100 with a sampling frequency of 1 kHz \cite{pintelon2012system}. The output was generated by a polynomial NARMAX system of degree $3$ (without mixed orders involving errors) and delays of $M_1$=$M_2$=$M_3$=$1$. In the first experiment, the noise was generated with a standard deviation of $0.02$, corresponding to a precision of $2500$. The coefficients $\theta$ were pseudo-randomly generated: $u_k$, $u_{k-1}$ and $y_{k-1}$ were assigned transfer function coefficients from a Butterworth filter with a cut-off frequency of 100 Hz and the coefficient for $e_{k-1}$ was assigned the value $0.1$. The remaining coefficients were sampled from uniform distributions centered at $0$ scaled by $0.01$.

VMP's prior precision parameters were set to $\alpha_0 = 10$ and $\beta_0 = 0.1$, corresponding to a mean of $100$ with a variance of $1000$. Note that this is not an informative prior as the true noise precision is $2500$. VMP's coefficients prior was set to be weakly informative, with $\mu_0 = {\bf 0}$ and $\Lambda_0 = I$. We generated 200 signal realizations and plot the average RMS along with the standard error of the mean (SEM) as ribbons.

\begin{figure}[htb]
    \includegraphics[width=.48\textwidth]{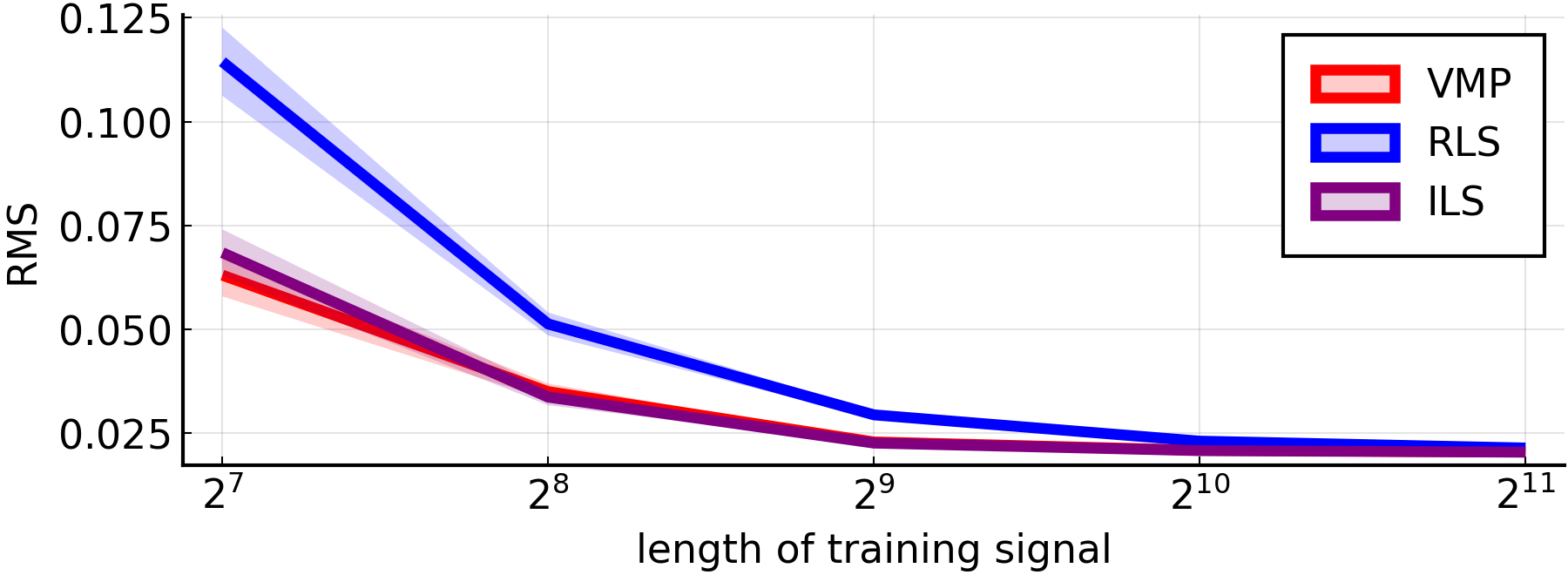}
    \caption{Simulation errors for a noise standard deviation of $0.02$. Average RMS (standard errors as ribbon) by length of training signal.}
    \label{fig:verif-simulation}
\end{figure}

\subsection{Results}
Figure \ref{fig:verif-simulation} shows the simulation errors of the three estimators as a function of the number of training samples. VMP outperforms RLS, especially for small sample sizes. This is due to the inclusion of the prior distributions and the regularizing effect of the parameter posterior on the predictions. VMP performs on par with ILS, which was trained offline. As sample size grows, the three estimators converge to the same level of performance.

\begin{figure}[htb]
    \includegraphics[width=.48\textwidth]{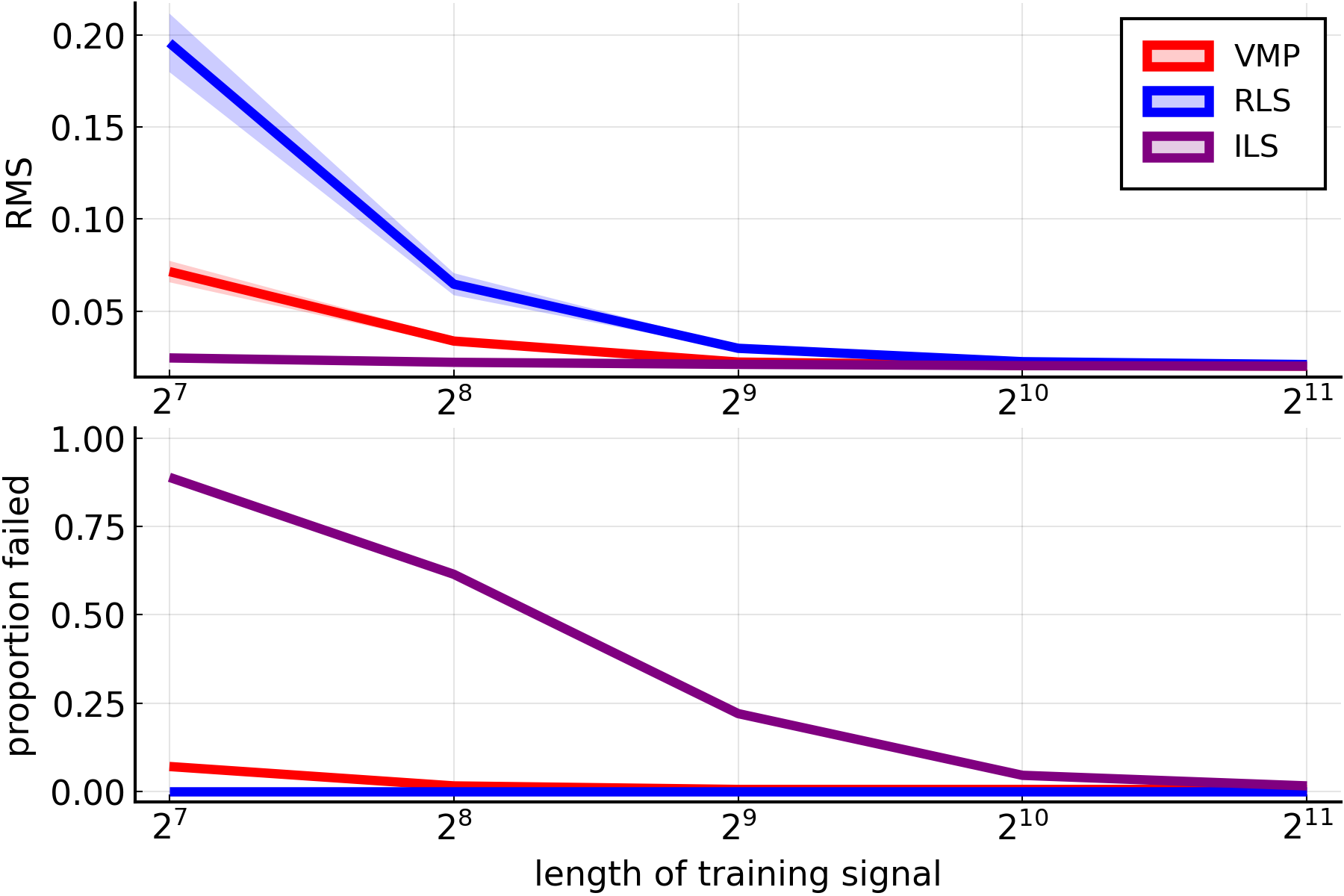}
    \caption{1-step ahead prediction errors for noise std. dev. $0.2$. (Top) Average RMS (standard errors as ribbons) by length of training signal. (Bottom) Proportion of experiments failed due to diverging parameter estimates.}
    \label{fig:verif-prediction}
\end{figure}
Figure \ref{fig:verif-prediction} (top) shows the 1-step ahead prediction errors as a function of the number of training samples. VMP still consistently outperforms RLS, but is no longer on par with ILS. Although ILS performs well, it also tends to diverge in small sample sizes: it would initially produce a prediction with just a slightly larger magnitude, but when the accompanying prediction error was incorporated back into the model, the next prediction would be even larger in magnitude. Figure \ref{fig:verif-prediction} (bottom) plots the proportion of failed experiments, i.e., those with diverging predictions, for all three estimators as a function of training signal length. ILS diverges less often as training signal length increases, with most of the failures having disappeared after 1024 samples. 

\begin{figure}[htb]
    \centering
    \includegraphics[width=.48\textwidth]{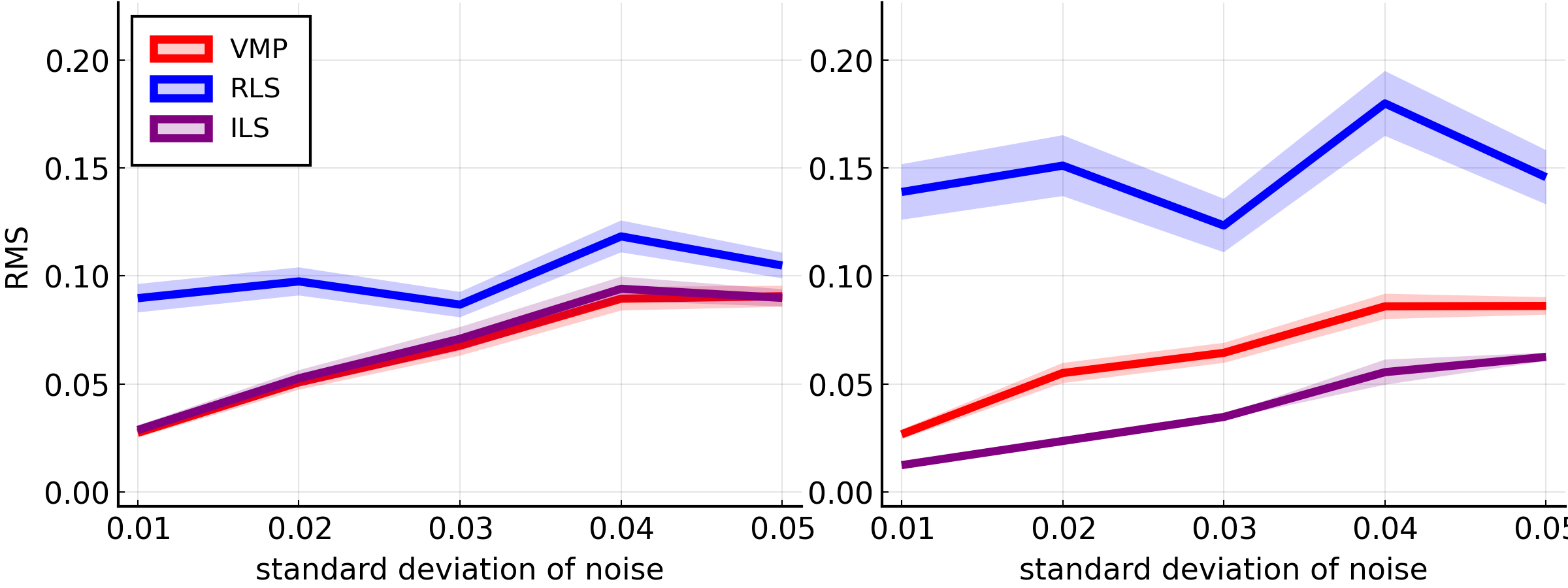}
    \vspace{-15px}
    \caption{Average RMS (standard errors as ribbons) as a function of system noise, for training signals of length $128$. Simulation (left) and 1-step ahead prediction (right).}
    \label{fig:sweep-stde}
\end{figure}

In our second experiment, we keep the training signal length fixed at $128$ and vary the standard deviation of the system's noise. Figure \ref{fig:sweep-stde} shows the average RMS of the three estimators, along with the standard errors, for simulation (left) and 1-step ahead prediction (right). VMP outperforms RLS for all levels of noise, but especially in the low noise levels. VMP performs on par with ILS during simulation but not during 1-step ahead prediction.

\section{Discussion} \label{sec:discussion}
Variational message passing is a modular procedure and can be automatized: tools such as ForneyLab.jl contain factor nodes in the form of standard parametric distributions, deterministic operations and common filters \cite{cox2018forneylab}. The advantage of modularity is that different models can be combined without the need for re-deriving parameter update equations \cite{senoz2020bayesian}. Among others, this allows for straightforward extensions towards hierarchical models and time-varying parameter estimates \cite{podusenko2021message}.
The main limitation of variational message passing is that it requires some form of conditional conjugacy in the prior distributions and recognition factors. Non-conjugate message passing is possible, but often comes at higher computational cost \cite{knowles2011non}.

\section{Conclusion}
We proposed a variational message passing algorithm for online system identification in polynomial NARMAX models. We show how to recursively update parameter posterior distributions and how to predict future outputs from given inputs. We demonstrated empirically that our estimator outperforms a recursive least-squares estimator and performs on par with an iterative least-squares estimator trained offline.

\section{Acknowledgements}
This work was partly financed by research programs ZERO (no. P15-06) and EDL (no. P16-25), funded by the Netherlands Organisation for Scientific Research (NWO).

\bibliographystyle{IEEEtran}
\bibliography{references}

\begin{thebibliography}{10}
\providecommand{\url}[1]{#1}
\csname url@samestyle\endcsname
\providecommand{\newblock}{\relax}
\providecommand{\bibinfo}[2]{#2}
\providecommand{\BIBentrySTDinterwordspacing}{\spaceskip=0pt\relax}
\providecommand{\BIBentryALTinterwordstretchfactor}{4}
\providecommand{\BIBentryALTinterwordspacing}{\spaceskip=\fontdimen2\font plus
\BIBentryALTinterwordstretchfactor\fontdimen3\font minus
  \fontdimen4\font\relax}
\providecommand{\BIBforeignlanguage}[2]{{%
\expandafter\ifx\csname l@#1\endcsname\relax
\typeout{** WARNING: IEEEtran.bst: No hyphenation pattern has been}%
\typeout{** loaded for the language `#1'. Using the pattern for}%
\typeout{** the default language instead.}%
\else
\language=\csname l@#1\endcsname
\fi
#2}}
\providecommand{\BIBdecl}{\relax}
\BIBdecl

\bibitem{billings2013nonlinear}
S.~A. Billings, \emph{Nonlinear system identification: {NARMAX} methods in the
  time, frequency, and spatio-temporal domains}.\hskip 1em plus 0.5em minus
  0.4em\relax John Wiley \& Sons, 2013.

\bibitem{murphy2012machine}
K.~P. Murphy, \emph{Machine learning: a probabilistic perspective}.\hskip 1em
  plus 0.5em minus 0.4em\relax MIT press, 2012.

\bibitem{sarkka2013bayesian}
S.~S{\"a}rkk{\"a}, \emph{Bayesian filtering and smoothing}.\hskip 1em plus
  0.5em minus 0.4em\relax Cambridge University Press, 2013.

\bibitem{peterka1981bayesian}
V.~Peterka, ``Bayesian approach to system identification,'' in \emph{Trends and
  Progress in System Identification}.\hskip 1em plus 0.5em minus 0.4em\relax
  Elsevier, 1981, pp. 239--304.

\bibitem{schon2015sequential}
T.~B. Sch{\"o}n, F.~Lindsten, J.~Dahlin, J.~W{\aa}gberg, C.~A. Naesseth,
  A.~Svensson, and L.~Dai, ``Sequential {Monte Carlo} methods for system
  identification,'' \emph{IFAC-PapersOnLine}, vol.~48, no.~28, pp. 775--786,
  2015.

\bibitem{blei2017variational}
D.~M. Blei, A.~Kucukelbir, and J.~D. McAuliffe, ``Variational inference: A
  review for statisticians,'' \emph{Journal of the American Statistical
  Association}, vol. 112, no. 518, pp. 859--877, 2017.

\bibitem{friston2008variational}
K.~J. Friston, N.~Trujillo-Barreto, and J.~Daunizeau, ``{DEM}: a variational
  treatment of dynamic systems,'' \emph{Neuroimage}, vol.~41, no.~3, pp.
  849--885, 2008.

\bibitem{daunizeau2009variational}
J.~Daunizeau, K.~J. Friston, and S.~J. Kiebel, ``Variational {Bayesian}
  identification and prediction of stochastic nonlinear dynamic causal
  models,'' \emph{{Physica D: Nonlinear Phenomena}}, vol. 238, no.~21, pp.
  2089--2118, 2009.

\bibitem{meera2020free}
A.~A. Meera and M.~Wisse, ``Free energy principle based state and input
  observer design for linear systems with colored noise,'' in \emph{American
  Control Conference}, 2020, pp. 5052--5058.

\bibitem{risuleo2017variational}
R.~S. Risuleo, G.~Bottegal, and H.~Hjalmarsson, ``Variational {Bayes}
  identification of acyclic dynamic networks,'' \emph{IFAC-PapersOnLine},
  vol.~50, no.~1, pp. 10\,556--10\,561, 2017.

\bibitem{hendriks2021deep}
J.~N. Hendriks, F.~K. Gustafsson, A.~H. Ribeiro, A.~G. Wills, and T.~B.
  Sch{\"o}n, ``Deep energy-based {NARX} models,'' \emph{IFAC-PapersOnLine},
  vol.~54, no.~7, pp. 505--510, 2021.

\bibitem{loeliger2004introduction}
H.-A. Loeliger, ``An introduction to factor graphs,'' \emph{IEEE Signal
  Processing Magazine}, vol.~21, no.~1, pp. 28--41, 2004.

\bibitem{korl2005factor}
S.~Korl, ``A factor graph approach to signal modelling, system identification
  and filtering,'' Ph.D. dissertation, ETH Zurich, 2005.

\bibitem{dauwels2007variational}
J.~Dauwels, ``On variational message passing on factor graphs,'' in \emph{IEEE
  International Symposium on Information Theory}, 2007, pp. 2546--2550.

\bibitem{fujimoto2016system}
K.~Fujimoto and Y.~Takaki, ``On system identification for {ARMAX} models based
  on the variational {Bayesian} method,'' in \emph{Conference on Decision and
  Control}.\hskip 1em plus 0.5em minus 0.4em\relax IEEE, 2016, pp. 1217--1222.

\bibitem{lu2014variational}
Y.~Lu, S.~Khatibisepehr, and B.~Huang, ``A variational {Bayesian} approach to
  identification of switched {ARX} models,'' in \emph{IEEE Conference on
  Decision and Control}, 2014, pp. 2542--2547.

\bibitem{kouw2020online}
W.~M. Kouw, ``Online system identification in a {Duffing} oscillator by free
  energy minimisation,'' in \emph{International Workshop on Active
  Inference}.\hskip 1em plus 0.5em minus 0.4em\relax Springer, 2020, pp.
  42--51.

\bibitem{jacobs2018sparse}
W.~R. Jacobs, T.~Baldacchino, T.~Dodd, and S.~R. Anderson, ``Sparse {Bayesian}
  nonlinear system identification using variational inference,'' \emph{IEEE
  Transactions on Automatic Control}, vol.~63, no.~12, pp. 4172--4187, 2018.

\bibitem{yedidia2005constructing}
J.~S. Yedidia, W.~T. Freeman, and Y.~Weiss, ``Constructing free-energy
  approximations and generalized belief propagation algorithms,'' \emph{IEEE
  Transactions on Information Theory}, vol.~51, pp. 2282--2312, 2005.

\bibitem{khandelwal2018simulation}
D.~Khandelwal, M.~Schoukens, and R.~T{\'o}th, ``On the simulation of polynomial
  {NARMAX} models,'' in \emph{IEEE Conference on Decision and Control}, 2018,
  pp. 1445--1450.

\bibitem{hayes2009statistical}
M.~H. Hayes, \emph{Statistical digital signal processing and modeling}.\hskip
  1em plus 0.5em minus 0.4em\relax John Wiley \& Sons, 2009.

\bibitem{pintelon2012system}
R.~Pintelon and J.~Schoukens, \emph{System identification: a frequency domain
  approach}.\hskip 1em plus 0.5em minus 0.4em\relax John Wiley \& Sons, 2012.

\bibitem{cox2018forneylab}
M.~Cox, T.~van~de Laar, and B.~de~Vries, ``Forneylab.jl: Fast and flexible
  automated inference through message passing in julia,'' in
  \emph{International Conference on Probabilistic Programming}, 2018.

\bibitem{senoz2020bayesian}
{\.I}.~\c{S}en{\"o}z, A.~Podusenko, W.~M. Kouw, and B.~de~Vries, ``Bayesian
  joint state and parameter tracking in autoregressive models,'' in
  \emph{Conference on Learning for Dynamics and Control}, 2020, pp. 1--10.

\bibitem{podusenko2021message}
A.~Podusenko, W.~M. Kouw, and B.~de~Vries, ``Message passing-based inference
  for time-varying autoregressive models,'' \emph{Entropy}, vol.~23, no.~6, p.
  683, 2021.

\bibitem{knowles2011non}
D.~Knowles and T.~Minka, ``Non-conjugate variational message passing for
  multinomial and binary regression,'' \emph{Advances in Neural Information
  Processing Systems}, vol.~24, pp. 1701--1709, 2011.

\end{thebibliography}

\end{document}